\documentclass[conference]{IEEEtran}
\IEEEoverridecommandlockouts
\usepackage{cite}
\usepackage{amsmath,amssymb,amsfonts}
\usepackage{algorithmic}
\usepackage{graphicx}
\usepackage{textcomp}
\usepackage{xcolor}

\usepackage{url}
\usepackage{multirow}

\def\BibTeX{{\rm B\kern-.05em{\sc i\kern-.025em b}\kern-.08em
    T\kern-.1667em\lower.7ex\hbox{E}\kern-.125emX}}
    
\begin{document}

\title{Incremental Training and Group Convolution Pruning for Runtime DNN Performance Scaling on Heterogeneous Embedded Platforms\\
}

\author{\IEEEauthorblockN{Lei Xun, Long Tran-Thanh, Bashir M Al-Hashimi, Geoff V. Merrett}
\IEEEauthorblockA{\textit{School of Electronics and Computer Science} \\
\textit{University of Southampton}\\
Southampton, UK \\
{\{lx2u16, ltt08r, bmah, gvm\}}@ecs.soton.ac.uk}

}

\IEEEoverridecommandlockouts
\IEEEpubid{\makebox[\columnwidth]{978-1-7281-5758-0/19/\$31.00 ©2019 IEEE\hfill} \hspace{\columnsep}\makebox[\columnwidth]{ }}

\maketitle

\IEEEpubidadjcol

\begin{abstract}
Inference for Deep Neural Networks is increasingly being executed locally on mobile and embedded platforms due to its advantages in latency, privacy and connectivity. Since modern System on Chips typically execute a combination of different and dynamic workloads concurrently, it is challenging to consistently meet inference time/energy budget at runtime because of the local computing resources available to the DNNs vary considerably. To address this challenge, a variety of dynamic DNNs were proposed. However, these works have significant memory overhead, limited runtime recoverable compression rate and narrow dynamic ranges of performance scaling. In this paper, we present a dynamic DNN using incremental training and group convolution pruning. The channels of the DNN convolution layer are divided into groups, which are then trained incrementally. At runtime, following groups can be pruned for inference time/energy reduction or added back for accuracy recovery without model retraining. In addition, we combine task mapping and Dynamic Voltage Frequency Scaling (DVFS) with our dynamic DNN to deliver finer trade-off between accuracy and time/power/energy over a wider dynamic range. We illustrate the approach by modifying AlexNet for the CIFAR10 image dataset and evaluate our work on two heterogeneous hardware platforms: Odroid XU3 (ARM big.LITTLE CPUs) and Nvidia Jetson Nano (CPU and GPU). Compared to the existing works, our approach can provide up to 2.36x (energy) and 2.73x (time) wider dynamic range with a 2.4x smaller memory footprint at the same compression rate. It achieved 10.6x (energy) and 41.6x (time) wider dynamic range by combining with task mapping and DVFS.\\
\end{abstract}


\section{Introduction}
In the last few years, Deep Neural Networks (DNNs)\cite{deeplearning} have gained lots of attention and been widely adopted in many computer vision tasks such as image classification\cite{alexnet}, object detection\cite{yolo} and face recognition\cite{deepface} due to their ability to deliver near or super-human accuracy. 

\begin{table}[htbp]
\begin{center}
\caption{Mean inference time for a single CIFAR10 image using AlexNet}
\begin{tabular}{|c|c|c|c|}
\cline{1-4} 
\textbf{Platform} & \textbf{Computing cores} & \textbf{Time (ms)} & \textbf{ Accuracy (\%)}\\
\hline
\multirow{4}{*}{Jetson Nano}& GPU (614MHz) & 7.21 & \multirow{8}{*}{71.2}  \\
& GPU (921MHz) & 4.88 &  \\
& A57 CPU (921MHz) & 70.4 &   \\
& A57 CPU (1.43GHz) & 46.8 &   \\
\cline{1-3} 
\multirow{4}{*}{Odroid XU3}& A15 CPU (200MHz) & 1020 &   \\
& A15 CPU (1.8GHz) & 117 &   \\
& A7 CPU (200MHz) & 1780 &   \\
& A7 CPU (1.3GHz) & 280 &   \\
\hline 
\end{tabular}
\label{tab1}
\end{center}
\end{table}

The execution of DNNs has two stages: training and inference. At the training stage, DNNs learn to perform a task. Millions of DNN parameters need to be adjusted during this process. Therefore training is usually executed on powerful desktop/server GPU(s). At the inference (also known as testing) stage, DNNs perform the task on unseen data using pre-trained parameters. Inference can also be executed on desktop/server GPU(s), but it is increasingly being executed locally on mobile and embedded platforms due to its advantages in latency, privacy and connectivity\cite{NetAdapt,reform}. 

The performance of inference can be defined using platform-dependent metrics such as execution time and energy consumption, and platform-independent metrics such as classification accuracy and confidence. As shown in Table \ref{tab1}, when the same image classification DNN is deployed on two different platforms, the inference time varies considerably, whereas the accuracy remains the same. 

Modern mobile and embedded System on Chips (SoCs) typically execute a combination of different and dynamic workloads concurrently on heterogeneous computing cores (e.g. CPU, GPU, NPU). The use of runtime resource management techniques (e.g. task mapping and Dynamic Voltage Frequency Scaling (DVFS)) make these SoCs highly efficient. It is challenging to consistently meet inference time/energy budget at runtime because of the local computing resources available to the DNNs vary considerably. For example, the target computing cores might be unavailable due to other applications running on them, or are available at a lower voltage/frequency level because of other computing cores execute in the same voltage/frequency domain, or fixed power/thermal budgets. To address this challenge, a variety of dynamic DNNs were proposed to cover the performance variance at runtime. The computation workload of dynamic DNNs can be scaled through model compression at runtime to meet the performance budget. However, these works have significant memory overhead, limited runtime recoverable compression rate (RRCR) and narrow dynamic ranges of performance scaling. RRCR is defined as the percentage of filters/channels can be pruned and added during runtime without model retraining.

In this paper, we present a dynamic DNN using incremental training and group convolution pruning. The channels of the DNN convolution layer are divided into groups, which are then trained incrementally. At runtime, following groups can be pruned for inference time/energy reduction or added back for accuracy recovery without model retraining. Compared to the existing works, our approach can provide up to 2.36x (energy) and 2.73x (time) wider dynamic range with a 2.4x smaller memory footprint at the same compression rate. Moreover, previous works did not consider the runtime resource management techniques on SoCs. We combine our dynamic DNN with task mapping and DVFS to deliver a finer trade-off over performance metrics, and up to 10.6x (energy) and 41.6x (time) wider dynamic ranges.

The contributions of this paper are:

\begin{itemize}
  \item Proposed an approach for building dynamic DNNs using group convolution with incremental training.
  \item The first implementation of dynamic DNN with task mapping and DVFS to achieve finer performance trade-offs and wider dynamic ranges of performance scaling than standalone dynamic DNN approaches.
\end{itemize}

\section{Related Work}

Modern SoCs typically execute a combination of different dynamic workloads concurrently, and hence the local resources available to the DNN vary considerably at runtime. Static DNN compression \cite{NetAdapt, he2018amc} generates one DNN for a given performance budget at a pre-defined hardware setting (e.g. computing core and voltage/frequency level). This raises a significant problem since the performance budgets cannot be met when the pre-defined hardware setting is unavailable at runtime. Multiple DNNs are needed to cover all hardware settings, which result in significant memory storage overhead. Furthermore, the switching activities of these DNNs at runtime may cause significant delay and energy consumption\cite{park2015big}. 

To address this problem, a variety of dynamic DNNs were proposed. Dynamic DNNs can be partially executed at runtime to meet the performance budget using available resources. Xu \textit{et al.} \cite{reform} proposed a two-step DNN compression scheme. At design time, static DNN compression similar to \cite{NetAdapt} is used to adapt DNN on target hardware. At runtime, filters in the static compressed DNN are pruned further to meet dynamic budgets, or are added back for accuracy recovery without model retraining. Although this approach allows DNNs to adapt to dynamic resource variance (e.g. cores, frequency), the limited RRCR leads to a limited dynamic range, at most 20\% of the filters are pruned for a 25\% time/energy reduction. Such a narrow dynamic range is not enough to cover the performance variance when DNN is mapped on computing cores with lower performance than the pre-defined one, and/or with lower voltage/frequency level as shown in Table \ref{tab1}. Therefore, this approach still has significant memory storage overhead since multiple DNNs are needed to cover all hardware settings in modern SoCs.

In order to achieve a greater RRCR and wider dynamic range, DNNs can be initially designed to support runtime trade-off. Tann \textit{et al.} \cite{tann2016runtime} proposed a dynamic DNN using channel-wise incremental training. Unlike regular training which trains all channels and layers at the same time, channel-wise incremental training trains part of channels of all layers at a time. For example, For a four-increment dynamic DNN, 25\% of the channels of all layers are trained first, then another 25\% of the channels are trained while incorporates the pre-trained and frozen 25\% of channels, and so on so forth. After the training is finished, four discrete DNN configurations with different size/accuracy/time/energy are generated and stored as a single model. At runtime, the DNN is partially or fully executed depending on performance budgets and available resources. However, because even a smallest DNN configuration (e.g. with 25\% of channels) is required to perform the complete task (i.e. should have enough capacity for all data without under-fitting), this approach requires using an oversized model. 

\begin{figure*}
\centerline{\includegraphics[width=\textwidth]{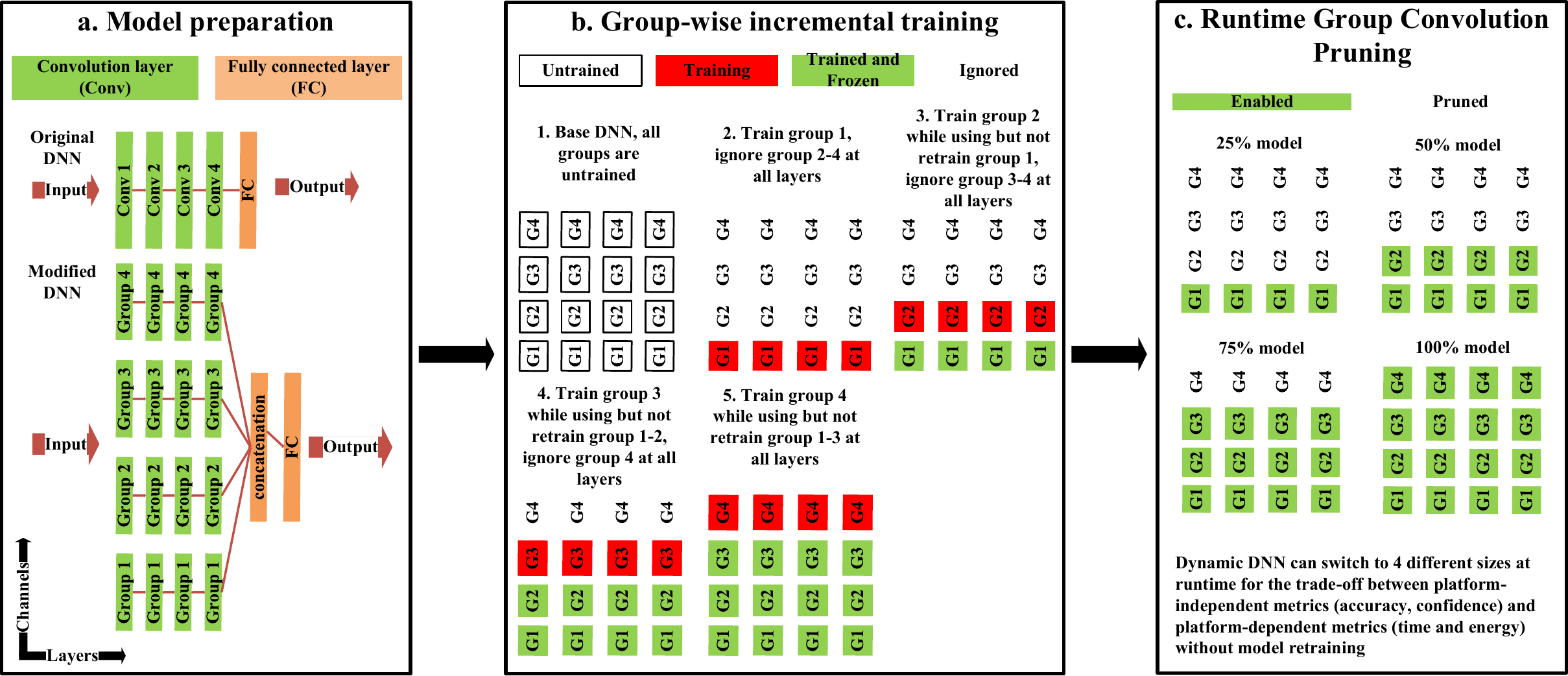}}
\caption{Our dynamic DNN using incremental training and group convolution pruning. The channels of the DNN convolution layer are (a) divided into groups, which are then (b) trained incrementally. At runtime, (c) following groups can be pruned for inference time/energy reduction or added back for accuracy recovery without model retraining.}
\label{fig1}
\end{figure*}

Although our work is similar to that in Tann \textit{et al.} \cite{tann2016runtime}, there are two main differences: 1) our work uses group convolution (Fig \ref{fig1}(a)). Each channel-wise incremental training is encapsulated in a group, and there are no connections between groups. This makes our design 2.4x smaller at the same compression rate. 2) we combine our dynamic DNN with task mapping and DVFS to deliver finer performance trade-offs over wider dynamic ranges.

\section{Incremental Training and Group Convolution Pruning}

This section introduces our approach of building a dynamic DNN from existing state-of-the-art DNN architectures. The building procedure has two steps: model preparation and group-wise incremental training. Once the model is trained, it can then be scaled at runtime to meet performance budgets using dynamically available resources without further model retraining. Moreover, the dynamic DNN is combined with task mapping and DVFS to deliver finer performance trade-offs over wider dynamic ranges.

\subsection{Model Preparation}

Our dynamic DNN approach is based on group convolution, which was first introduced in AlexNet\cite{alexnet}. AlexNet has two groups that are deployed on two GPUs separately due to limited GPU memory at that time. DNNs with group convolution are smaller and faster than their original configuration, as the dense connections between groups are disconnected, the model can have significantly fewer parameters. Group convolution has been used as the foundation of many state-of-the-art DNN architectures \cite{zhang2018shufflenet,xie2017aggregated}. 

The concept of our model preparation approach is illustrated in Fig \ref{fig1}(a). The channels of the DNN convolution layers are divided into groups which are then concatenated before the fully connected layer. Each group has its own activation layer (e.g. rectified linear unit (ReLU)), and there are no connections between groups. For DNNs that initially have branches\cite{inception} or residual connections\cite{resnet}, the model preparation can be seen as capturing the topology of the model as a building block which is then copied and concatenated, and the size of each building block is also reduced accordingly. This is equivalent to group convolution as shown in Xie \textit{et al.}\cite{xie2017aggregated}. However, unlike previous works which train all groups concurrently\cite{zhang2018shufflenet,xie2017aggregated}, our work trains these groups incrementally.

\subsection{Group-wise Incremental Training}

Each group tends to learn very different features due to less feature sharing between groups\cite{alexnet}. Based on this observation, incremental training\cite{tann2016runtime}  is incorporated in our work, and we use four increments in this paper to illustrate our approach. As shown in Fig \ref{fig1}(b-1), a base DNN is obtained from model preparation step (Fig \ref{fig1}(a)) and all of its groups are untrained. The training in our approach has four steps:
\begin{itemize}
  \item \textbf{Step 1:} Train group 1 of all layers, ignore all other groups through initialising all their parameters with zero value so that they do not affect the output (Fig \ref{fig1}(b-2)).
  \item \textbf{Step 2:} Train group 2 of all layers while incorporate pre-trained group 1, ignore group 3-4 (Fig \ref{fig1}(b-3)).
  \item \textbf{Step 3:} Train group 3 of all layers while incorporate pre-trained group 1-2, ignore group 4 (Fig \ref{fig1}(b-4)).
  \item \textbf{Step 4:} Train group 4 of all layers while incorporate pre-trained group 1-3 (Fig \ref{fig1}(b-5)).
\end{itemize}
Each training steps generate a model that uses a different number of groups. Therefore these four models have different accuracy, confidence and computation requirements. Unlike four separated models, these models can be seen as four different DNN configurations within a single model. In this paper, we refer them as the 25\%, 50\%, 75\% and 100\% models as shown in Fig \ref{fig1}(c). All parameters are initialised randomly during training except those are in the ignored groups. The pre-trained groups are frozen so they cannot be changed during later training. The learning rate of the FC layer is reduced with every increment so that following groups have less impact on the 25\% model. This helps to make sure later training only add new knowledge to the previous knowledge rather than delete them. Training steps 3-5 are executed multiple times until a target accuracy improvement is met.

\subsection{Runtime Group Convolution Pruning}

Once all incremental training is finished, the model can be pruned progressively, one group at a time, as shown in Fig \ref{fig1}(c). For example, 25\% model uses only using one group of DNN parameters. Therefore it is the least accurate model but requires minimum computation. The 100\% model is the full model which is the most accurate and computationally expensive model. At runtime, the dynamic DNN can switch between these model configurations to explore the trade-off between platform-independent metrics (accuracy and confidence) and platform-dependent metrics (time and energy) without requiring model retraining.

\subsection{Task mapping and DVFS}

One disadvantage of incremental training is that the generated DNN configurations have sparse trade-offs. Since our approach is not tied to specific hardware, it can be combined with task mapping and DVFS to deliver finer performance trade-offs. \\

The effectiveness of our incremental training is validated through accuracy and confidence tests over four model configurations. Furthermore, runtime group convolution pruning is deployed on two heterogeneous embedded platforms and validated through empirical measurements of time/energy. More details are covered in the next section.

\section{Empirical validation}\label{val}

\subsection{Dynamic DNN Implementation Details}\label{imp}

We illustrate our approach using AlexNet\cite{alexnet} for the CIFAR10 image classification dataset\cite{cifar} with the Caffe framework\cite{jia2014caffe}. AlexNet is originally designed for the ILSVRC 2012 dataset\cite{ilsvrc} which contains around 1.3 million training images and 50,000 test images (image size 256*256*3) over 1000 image classes, whereas CIFAR10 only contains 50,000 training images and 10,000 test images (image size 32*32*3) over 10 image classes. Using AlexNet for CIFAR10 directly is unnecessary, and will most likely result in overfitting. Therefore, in our design, the layer size in AlexNet is reduced, but the topology of AlexNet is kept (except the reduced number of FC layers to fit CIFAR10). The detailed architecture is listed in Table \ref{tab2}. The model is trained incrementally, and around 50 epochs at each training step. In addition, intermediate models are saved during each training step. In training step 2, the model with the highest validation accuracy is selected as seed for the next increment. However, in training steps 3-5, a seed model is selected only when a target accuracy improvement is met. If no intermediate models meet the target, this step is then repeated.

\begin{table}[tbp]
\begin{center}
\caption{Modified AlexNet for CIFAR10}
\begin{tabular}{|c|c|c|}
\cline{1-3} 
\textbf{Layer} & \textbf{Configuration} & \textbf{Output size}\\
\hline
Input & N/A & [32*32*3] \\
\hline
Conv1 + ReLU & kernel=3, stride=1, pad=0 & [30*30*16]*4 \\
\hline
Norm1 & Local=5, alpha=0.0001, beta=0.75 & [30*30*16]*4 \\
\hline
MaxPool1 & kernel=4, stride=1 & [27*27*16]*4\\
\hline
Conv2 + ReLU & kernel=5, stride=1, pad=2 & [27*27*16]*4 \\
\hline
Norm2 & Local=5, alpha=0.0001, beta=0.75 & [27*27*16]*4 \\
\hline
MaxPool2 & kernel=3, stride=2 & [13*13*16]*4\\
\hline
Conv3 + ReLU & kernel=3, stride=1, pad=1 & [13*13*16]*4 \\
\hline
Conv4 + ReLU & kernel=3, stride=1, pad=1 & [13*13*16]*4 \\
\hline
Conv5 + ReLU & kernel=3, stride=1, pad=1 & [13*13*16]*4 \\
\hline
MaxPool5 & kernel=3, stride=2 & [6*6*16]*4\\
\hline
Concatention & N/A & [6*6*64]\\
\hline
FC6 & N/A & 10\\
\hline
Softmax & N/A & 10\\
\hline 
\end{tabular}
\label{tab2}
\end{center}
\end{table}

\subsection{Experimental Setup}

Our model is validated on two heterogeneous embedded platforms: the Nvidia Jetson Nano and Odroid XU3. On the Jetson Nano, the ARM A57 CPU and 128-core Maxwell GPU are used; both are configured with two different frequency levels. On the XU3, the ARM A15 and A7 CPUs are used with 17 and 12 different frequency levels respectively. All measurements using CPUs use only a single core because the program is single-threaded, and the batch size for GPUs is 1.

\begin{figure}[tbp]
\centerline{\includegraphics[width=\columnwidth]{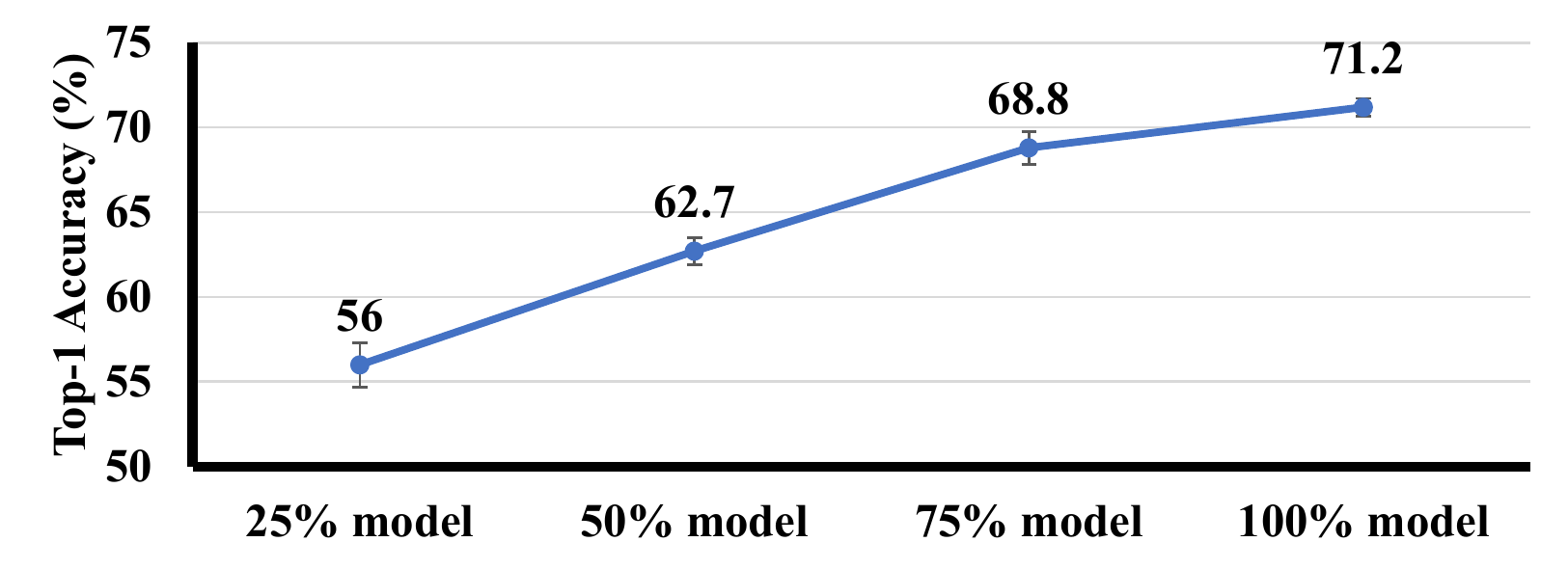}}
\caption{Top-1 image classification accuracy on 10,000 CIFAR10 validation images. Our Dynamic DNN has four different model configurations which have four different accuracies. At runtime, dynamic DNN can switch to smaller configuration for time/energy reduction with accuracy loss, or switch back to larger models for the accuracy recovery once more computing resources become available.}
\label{fig2}
\end{figure}

\begin{figure}[tbp]
\centerline{\includegraphics[width=\columnwidth]{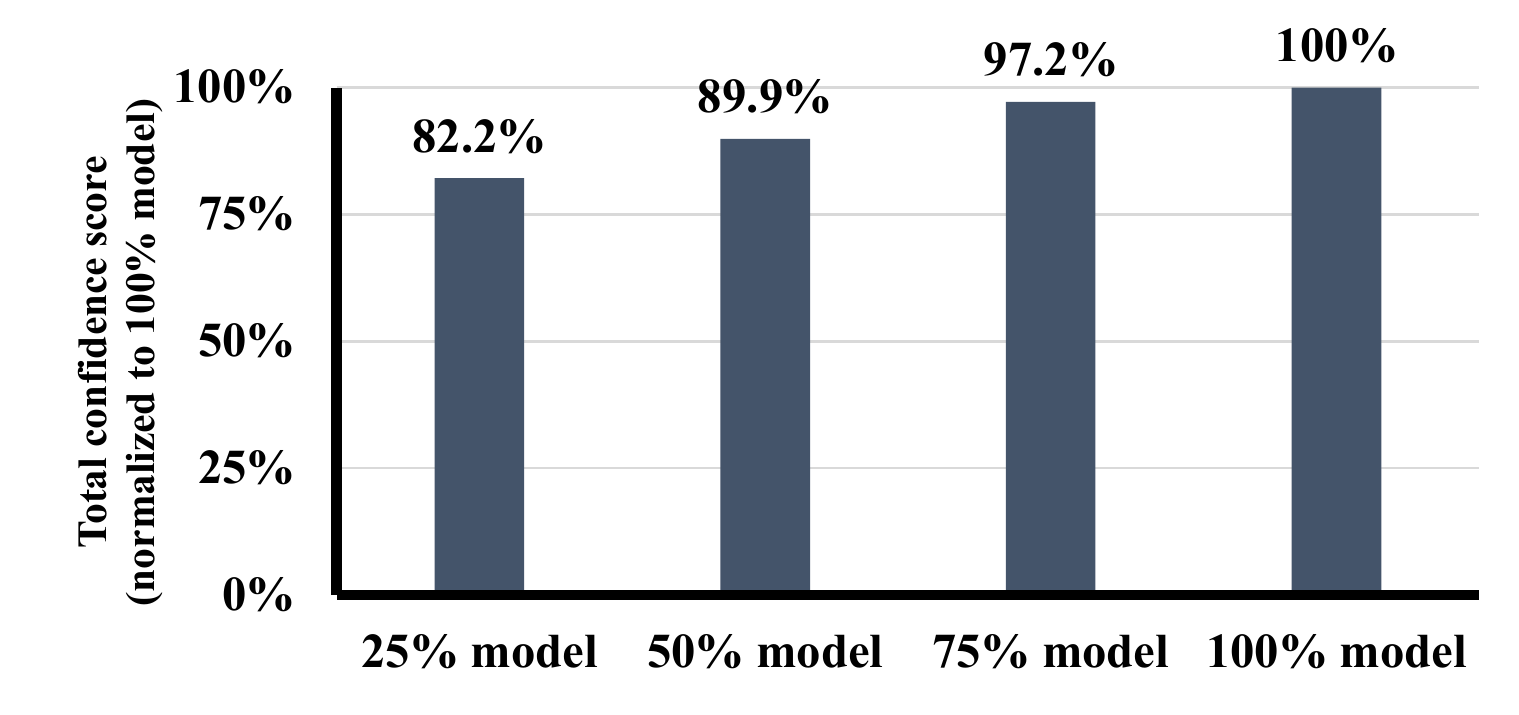}}
\caption{The total confidence score of the correct DNN output over 10,000 CIFAR10 validation images. The value is normalised to the 100\% model. The confidence is improved when more groups are added to dynamic DNN, this indicates different feature filters are learnt in following groups.}
\label{fig3}
\end{figure}

\begin{figure*}
\centerline{\includegraphics[width=\textwidth]{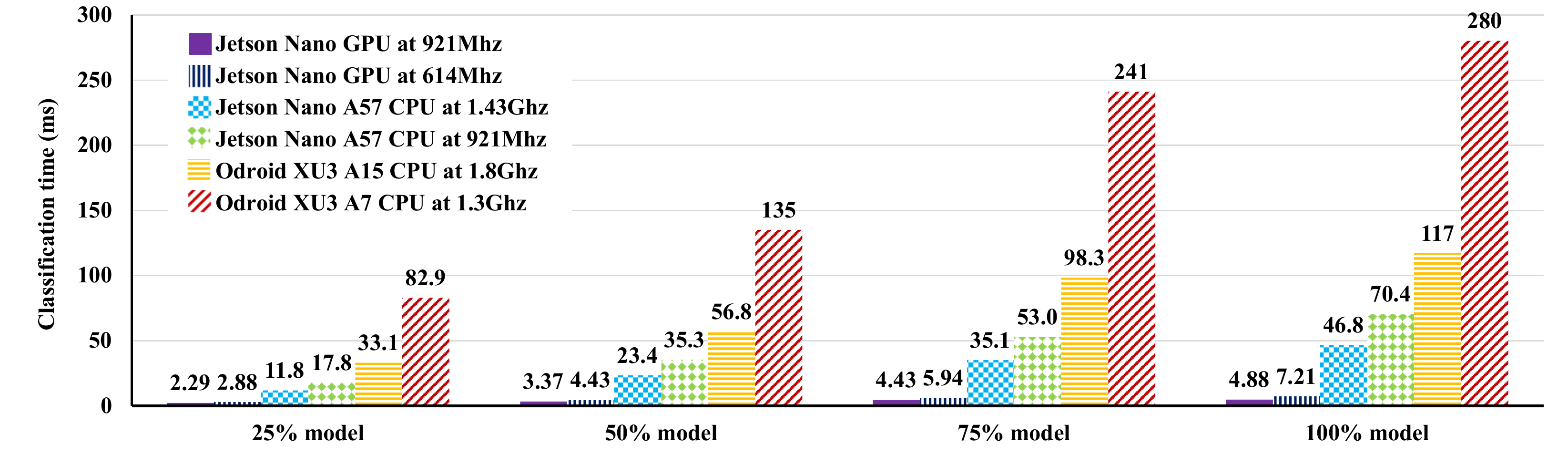}}
\caption{Inference time on four heterogeneous cores of two hardware platforms. Our group convolution pruning is fully compatible with both CPU and GPU.}
\label{fig4}
\end{figure*} 

\subsection{Inference Top-1 Accuracy and Confidence}\label{acc}

To validate our work, accuracy and confidence are measured over four different DNN configurations. Accuracy measures how accurate a model is for a given validation image dataset, and it is defined as:
 
\[
    \text{Accuracy} = \frac{\text{Number of correctly classified images}}{\text{Total number of images}}
\]

Each output of the DNN has a value, and the confidence score of an output is defined as:

\[
    \text{Confidence score} = \frac{\text{Value of the output}}{\text{Sum of the value of all outputs}}
\]

The confidence score of an output is higher when more image filters are matched with the patterns of the image class in input images and the feature maps at following layers. Top-1 accuracy is defined as the percentage of images that are classified correctly (also with the highest confidence score) in the entire dataset.

Our dynamic DNN has four different model configurations which have four different accuracies, as shown in Fig \ref{fig2}. The error bar shows the variance over 10 image classes of CIFAR10. Although there is accuracy loss during configuration switching, these losses are fully recoverable without any retraining. Fig \ref{fig3} shows the improvement in total confidence score over 10,000 CIFAR10 validation images. The improvement is normalised to the 100\% model. Confidence is improved when more groups are added to the dynamic network. This matches our expectation since different groups tend to learn different feature filters, and the confidence score is improved when more filters are matched. At runtime, the dynamic DNN can switch to a smaller configuration using group convolution pruning for time/energy reduction with accuracy loss, or switch back to larger models for the accuracy recovery once more computing resources become available. 

\subsection{Adapting to Different Hardware Platforms and Heterogeneous Cores}\label{GCP}

In this experiment, runtime group convolution pruning and its dynamic range are tested by deploying the dynamic DNN on four heterogeneous cores of two hardware platforms. As shown in Fig \ref{fig4}, the inference time is platform-dependent, therefore using a single model to achieve consistent time budget is hard since different platforms/cores have considerably different computing capabilities. Our dynamic DNN has up to 2.5x and 4x dynamic range on GPU and CPU, respectively. These dynamic ranges support the same model to be deployed on different computing cores across hardware platforms while achieving the same time budget. For example, for a time budget of 33ms (30 fps), the dynamic DNN can be deployed on Odroid XU3 A15 CPU using the 25\% model, or on the Jetson Nano GPU using the 100\% model at a lower frequency setting or A57 CPU using a 50\% model if the GPU is unavailable.

\subsection{Combining Dynamic DNN with Task mapping and DVFS}\label{DVFS}

\begin{figure*}
\centerline{\includegraphics[width=\textwidth]{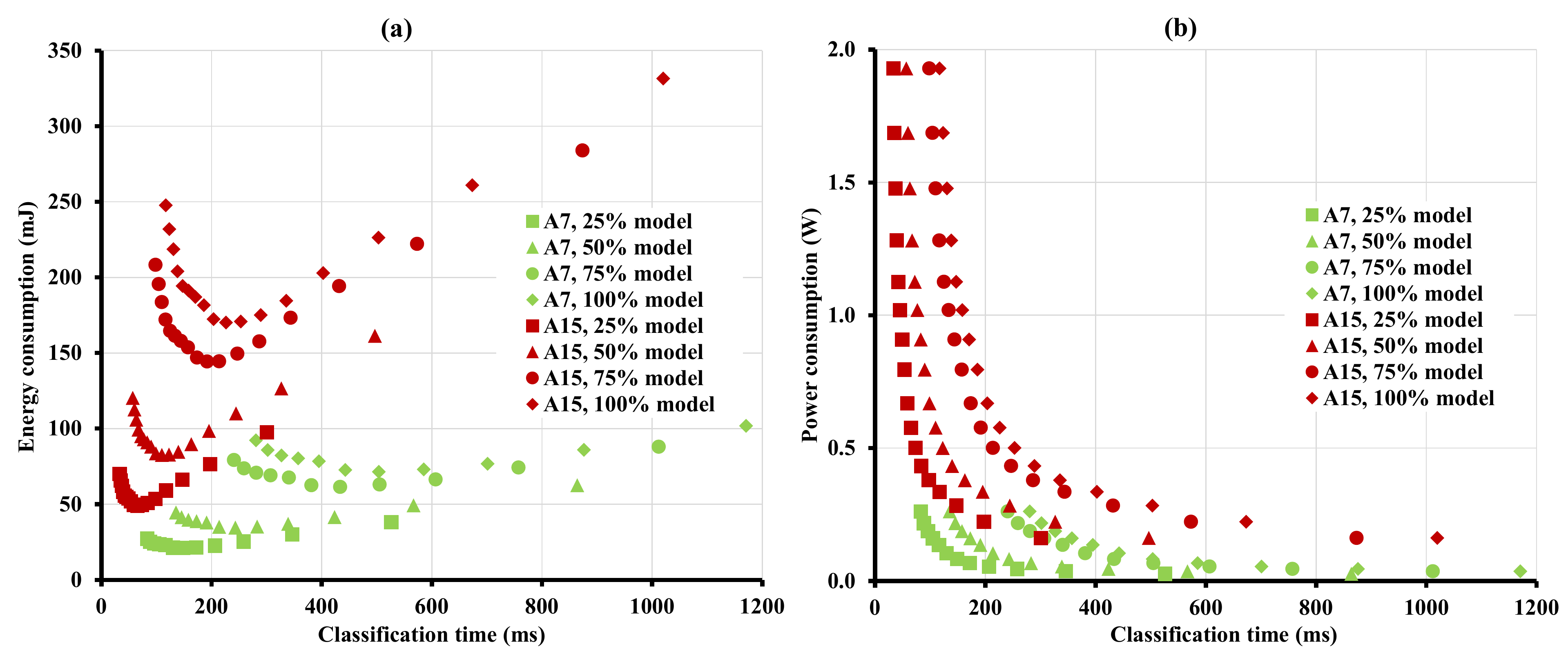}}
\caption{Dynamic DNN (different symbols) is combined with task mapping (different colours) and DVFS (different points). Different configurations are used for different runtime energy (a), power (b) and time budget. A standalone dynamic DNN  with four increments can only provide four trade-off points over a limited dynamic range. With the combination of task mapping and DVFS, the trade-offs are finer and the dynamic ranges are wider.}
\label{fig5}
\end{figure*}

This experiment further explores the combination of our dynamic DNN, task mapping and DVFS. The proposed dynamic DNN is deployed on both A15 and A7 CPUs of Odroid XU3 with 17 and 12 different frequency levels respectively. At runtime, combinations of these three adjustable "knobs" can be applied to meet considerably different energy (Fig \ref{fig5}a), power (Fig \ref{fig5}b) and time budgets.  A standalone dynamic DNN  with four increments can only provide four trade-off points over a limited dynamic range. With the combination of task mapping and DVFS, the trade-offs are finer and the dynamic ranges are wider.
 
\begin{table}[bp]
\begin{center}
\caption{Comparison with existing works}
\begin{tabular}{|c|c|c|c|}
\cline{1-4} 
\textbf{DNN} & \textbf{RRCR$^{\mathrm{1}}$} & \textbf{Dynamic range} & \textbf{Model size (KB)}\\
\hline
\multirow{2}{*}{Xu \textit{et al.}\cite{reform}}  & \multirow{2}{*}{20\%} & 0.25x (time) &  \multirow{2}{*}{773.9*N$^{\mathrm{3}}$} \\
&  & 0.25x (energy) & \\
\hline
\multirow{2}{*}{Tann \textit{et al.} \cite{tann2016runtime}} & \multirow{2}{*}{75\%} & 1.29x (time) & \multirow{2}{*}{773.9} \\
&  & 1.49x (energy) & \\
\hline 
\multirow{2}{*}{Proposed w/o D\&T$^{\mathrm{2}}$}& \multirow{2}{*}{75\%} & 3.53x (time) &  \multirow{2}{*}{318.4} \\
&  & 3.53x (energy) & \\
\hline 
\multirow{2}{*}{Proposed with DVFS} & \multirow{2}{*}{75\%} & 30.8x (time) &  \multirow{2}{*}{318.4} \\
&  & 6.76x (energy) & \\
\hline 
\multirow{2}{*}{Proposed with D\&T} & \multirow{2}{*}{75\%} & 53.7x (time) &  \multirow{2}{*}{318.4} \\
&  & 15.73x (energy) & \\
\hline 
\multicolumn{4}{l}{$^{\mathrm{1}}$RRCR is runtime recoverable compression rate.} \\
\multicolumn{4}{l}{$^{\mathrm{2}}$D\&T is DVFS and task mapping.} \\
\multicolumn{4}{l}{$^{\mathrm{3}}$N is set to 29 to cover all hardware settings on Odroid XU3.}
\end{tabular}
\label{tab3}
\end{center}
\end{table}

\subsection{Comparison against State-of-the-art Works}
We compare our work with existing works, and the results are shown in Table \ref{tab3}. Our approach provides large RRCR since the DNN is trained incrementally to support runtime pruning. Compared to Tann \textit{et al.} \cite{tann2016runtime}, our work has the same 75\% RRCR due to the same four increments DNN design. Our approach can provide up to 2.36x (energy) and 2.73x (time) wider dynamic ranges with a 2.4x smaller memory footprint due to the use of group convolution. Moreover, our work can achieve 10.6x (energy) and 41.6x (time) wider dynamic ranges by combining with task mapping and DVFS. Filter pruning based compression \cite{reform} has limited RRCR since most filters are needed to guarantee the DNN is still able to classify all image classes. Aggressive pruning can result in significant accuracy loss in some image classes where the features cannot be detected due to the missing filters. In addition, this approach optimises DNN for a pre-defined hardware setting (e.g. core, frequency). Therefore, it cannot be applied with task mapping and DVFS, and has significant memory storage overhead since multiple models are needed to cover all hardware settings. 

\section{Conclusion}

In this paper, we proposed a dynamic DNN using incremental training and group convolution pruning. The channels of the DNN convolution layer are divided into groups, which are then trained incrementally. At runtime, following groups can be pruned for inference time/energy reduction or added back for accuracy recovery without model retraining. Compared to state-of-the-art, our approach can provide up to 2.36x (energy) and 2.73x (time) wider dynamic ranges with a 2.4x smaller memory footprint at the same compression rate. Moreover, our work can achieve 10.6x (energy) and 41.6x (time) wider dynamic ranges by combining with task mapping and DVFS.

\section{Acknowledgement}

This work was supported in part by the Engineering and Physical Sciences Research Council (EPSRC) under Grant EP/S030069/1. Experimental data can be found at DOI: 10.5258/SOTON/D1245.



\end{document}